\def\year{2020}\relax
\documentclass[letterpaper]{article} 
\usepackage{aaai20}  
\usepackage{times}  
\usepackage{helvet} 
\usepackage{courier}  
\usepackage[hyphens]{url}  
\usepackage{graphicx} 
\usepackage{graphicx}  
\usepackage[utf8]{inputenc} 
\usepackage[T1]{fontenc}    
\usepackage{hyperref}       
\usepackage{url}            
\usepackage{booktabs}       
\usepackage{amsfonts}       
\usepackage{nicefrac}       
\usepackage{microtype}      

\usepackage{calc}

\usepackage{array} 
\usepackage{tabularx}
\usepackage{booktabs}
\usepackage{expl3}[2018/10/31]
 
\usepackage{comment}

 \usepackage{booktabs,makecell}  
\usepackage{amsmath,graphicx}
\usepackage{threeparttable}
\usepackage{booktabs}
\usepackage{amsmath,graphicx}
\usepackage{url}
\usepackage{psfrag}
\usepackage{amssymb}
\usepackage{amsbsy}
\usepackage{graphicx}
\usepackage{array}
\usepackage{multirow}
\usepackage{bm}
\usepackage{verbatim}
\usepackage{relsize}
\usepackage{algorithm}
\usepackage{algpseudocode}
\usepackage{tabularx}
 \usepackage{verbatim}
\usepackage{bm}
\usepackage{lipsum}
\usepackage{tabularx}
\usepackage{mwe} 
\usepackage[utf8]{inputenc} 
\usepackage[T1]{fontenc}    
\usepackage{hyperref}       
\usepackage{url}            
\usepackage{booktabs}       
\usepackage{nicefrac}       
\usepackage{microtype}      
\usepackage{times}
\usepackage{epsfig}
\usepackage{graphicx}
\usepackage{amsmath}
\usepackage{amsthm}
\usepackage{mathrsfs} 
\usepackage{amssymb}
\usepackage{wrapfig}
\usepackage{lscape}
\usepackage{rotating}

\usepackage{wrapfig}

\newcommand{\Xt}[1]{\bm{X}^{(#1)}}

\newtheorem{thm}{Theorem}[section]

\newcommand{\Task}[1]{\mathcal{Z}^{(#1)}}

\def\w{\mathbf{w}}

\def\w{\mathbf{w}}

\newcommand{\Yt}[1]{\bm{Y}^{(#1)}}

\theoremstyle{plain}

\theoremstyle{definition}

\def\w{\mathbf{w}}

\def\w{\mathbf{w}}

\usepackage{color}

\usepackage{caption}
\usepackage{subcaption}
\usepackage{wrapfig}
\usepackage{tabularx}
\usepackage{placeins}
\usepackage{dblfloatfix}

\title{Generative Continual Concept Learning}
\author{%
  Mohammad~Rostami\\
  University of Pennsylvania\\
  Philadelphia, PA 19104 \\
  \texttt{mrostami@seas.upenn.com} \\
   \And
   Soheil Kolouri \\
 HRL Laboratories, LLC\\
  Malibu, CA 90265 \\
   \texttt{skolouri@hrl.com} \\
   \And
   James McClelland \\
   Stanford University \\
   Stanford, CA 94305\\
   \texttt{mcclelland@stanford.edu} \\
   \And
    Praveen Pilly \\
    HRL Laboratories, LLC\\
    Malibu, CA 90265 \\
   \texttt{pkpilly@hrl.com} \\
}
 
\begin{document}

\maketitle

\begin{abstract}
After learning a concept, humans are also able to continually generalize their learned concepts to new domains by observing only a few labeled instances without any interference with the past learned knowledge. In contrast, learning concepts efficiently in a continual learning setting remains an open challenge for current Artificial Intelligence algorithms as persistent model retraining is necessary. Inspired by the Parallel Distributed Processing learning and the Complementary Learning Systems theories, we develop a computational model that is able to expand its previously learned concepts efficiently to new domains using a few labeled samples. We couple the new form of a concept to its past learned forms in an embedding space for effective continual learning. Doing so, a generative distribution is learned such that it is shared across the tasks in the embedding space and models the abstract concepts. This procedure enables the model to generate pseudo-data points to replay the past experience to tackle catastrophic forgetting. 
 
\end{abstract}

\section{Introduction}
An important ability of humans is to continually build and update abstract concepts. Humans  develop and learn abstract concepts to characterize and communicate their perception and ideas~\cite{lake2015human}. These concepts often  are evolved and expanded efficiently as more experience about new domains is gained. Consider for example, the concept of the printed character ``4''. This concept is often taught   to represent the ``natural number four''   in the mother tongue of elementary school students, e.g., English. Upon learning this concept, humans can efficiently expand it by observing only a few samples from other related   domains, e.g., variety of hand written digits or printed digits in other secondary
languages.
Despite remarkable progress in Artificial intelligence (AI) over the past decade, learning concepts efficiently in a way similar to humans remains an unsolved challenge for AI.
This  is because the exceptional   progress of AI is mostly
driven by re-emergence of deep neural networks. Since deep networks are trained in an end-to-end supervised learning setting, access to labeled data is necessary for learning any new distribution. 
For this reason and despite emergence of behaviors similar to the nervous system in deep nets~\cite{morgenstern2014properties}, adapting a deep neural network to learn a concept in a new domain usually requires model retraining from scratch which is conditioned on the availability of   a large number of 
  labeled samples in the new domain. Moreover, training deep networks  in a continual learning setting   is challenging due to the phenomenon  of ``catastrophic forgetting''~\cite{french1999catastrophic}. When a network is trained on 
  sequential   tasks, the new learned knowledge usually  interferes with past learned knowledge, causing     forgetting   what has been learned before. 

In this paper, we develop a computational model that is able to expand and generalize  learned concepts efficiently to new domains using a few labeled data from the new domains. We rely on Parallel Distributed Processing (PDP) paradigm~\cite{mcclelland1986parallel} for this purpose.
Work on semantic cognition within the PDP 
framework  hypothesizes that abstract semantic concepts are formed in higher level layers of the nervous system~\cite{mcclelland2003parallel,saxe2019mathematical}. We model this hypothesis by assuming that  the data points are mapped into an embedding space, which captures existing concepts.
To prevent catastrophic forgetting, we rely on the Complementary Learning Systems (CLS) theory~\cite{mcclelland1995there}. CLS theory hypothesizes that continual lifelong learning ability of the nervous system is a result of  a dual long- and short-term memory system.  The hippocampus acts as short-term memory and   encodes recent experiences that are used to consolidate the knowledge in the neocortex as long-term memory through offline experience replays during sleep~\cite{diek2010}. This suggests that if we  store suitable samples from past domains in a memory buffer, like in the neocortex, these samples can be replayed along  with current task samples from recent-memory hippocampal storage to train the base model jointly on the past and the current experiences to  tackle catastrophic forgetting.  

More specifically, we model the latent  embedding space via responses of a hidden layer in a deep neural network. Our idea is to stabilize and consolidate the data distribution in this space, where domain-independent abstract concepts are encoded. Doing so,  new forms of concepts can be learned efficiently by coupling them to their past learned forms in the embedding space. Data representations in this embedding space can be considered as neocortical representations in the brain, where the learned abstract concepts are captured. We model concept learning in a sequential task learning framework, where learning concepts in each new domain is considered to be a task. To generalize the learned concepts without forgetting, we use an autoencoder as the base network to benefit from efficient coding ability of deep autoencoders and model the embedding space as the middle layer of the autoencoder. This will also  make our model generative, which can be used to implement the offline memory replay process in the sleeping brain~\cite{rasch2013}. To this end,  we fit a parametric multi-modal distribution to the training data representations in the embedding space. The drawn points from this distribution can be used to generate  pseudo-data points through the decoder network for experience replay to prevent catastrophic forgetting. We demonstrate that this learning procedure enables the base model to generalize its learned concepts to new domains using a few labeled samples.

\section{Related Work}

Lake et al.~\cite{lake2015human} modeled human concept learning within a ``Bayesian probabilistic learning'' (BPL) paradigm. They present BPL as an alternative for deep learning to mimic the learning ability of humans as  these models require considerably less amount of training data.   The concepts are 
  represented  as probabilistic programs that can generate additional instances of a concept given a few samples of that concept.  However, the proposed algorithm in Lake et al.~\cite{lake2015human}, requires human supervision and domain knowledge to tell the algorithm how the real-world concepts are generated. This approach seems feasible for the recognition task that they have designed to test their idea, but it does not scale to other more challenging concept learning problems.  
   Our framework similarly relies on a generative model that can produce pseudo-samples of the learned concepts, but we follow an end-to-end deep learning scheme that automatically encodes concepts in the hidden layer of the network with minimal human supervision requirement. Our approach can be applied to a broader range of problems. The price is that we rely on data to train the model, but only a few data points are labeled. This is similar to humans with respect to how they too need practice to generate samples of a concept when they do not have domain knowledge~\cite{longcamp2005influence}.  
   This generative strategy has been used in the Machine Learning (ML) literature to address ``few-shot learning'' (FSL)~\cite{snell2017prototypical,motiian2017few}.
 The goal of FSL is to adapt a model that is trained on a source domain with sufficient labeled data to generalize well on a \textit{related} target domain with a few target labeled data points.  In our work, the domains are different but also are related in that they share similar concepts.

Most FSL algorithms consider only one source and one target domain, which are learned jointly. Moreover, the main goal is  to learn the target task. In contrast, we consider a continual learning setting in which the domain-specific tasks arrive sequentially. Hence,  catastrophic forgetting becomes a major challenge. An effective approach to tackle catastrophic forgetting is to use experience replay~\cite{mccloskey1989catastrophic,robins1995catastrophic}. Experience replay addresses catastrophic forgetting  via storing and replaying data points of past learned tasks continually. Consequently, the model retains the probability distributions of the past learned tasks.  To avoid requiring a memory buffer to store past task samples,  generative models have been used to produce pseudo-data points for past tasks. To this end, generative adversarial learning can be used to match the cumulative distribution of the past tasks with the current task distribution to allow for generating pseudo-data points for experience replay~\cite{shin2017continual}. Similarly,  autoencoder structure can also be used to generate pseudo-data points~\cite{parisi2019continual,rostami2019Complementary}. Building upon our prior work~\cite{rostami2019Complementary}, we develop a new method for generative experience replay to tackle catastrophic forgetting. Although prior works require access to labeled data for all the sequential tasks for experience replay, we demonstrate that experience replay is feasible even in the setting where only the initial task has labeled data. Our contribution is to combine ideas of few-shot learning with generative experience replay to develop a framework that can continually update and generalize learned concepts when new domains are encountered in a lifelong learning setting. We couple the distributions of the tasks in the middle layer of an autoencoder and use the shared distribution to expand concepts using a few labeled data points without forgetting.
   
\section{Problem Statement and the Proposed Solution}

\begin{figure}[t!]
    \centering
    \includegraphics[width= \linewidth]{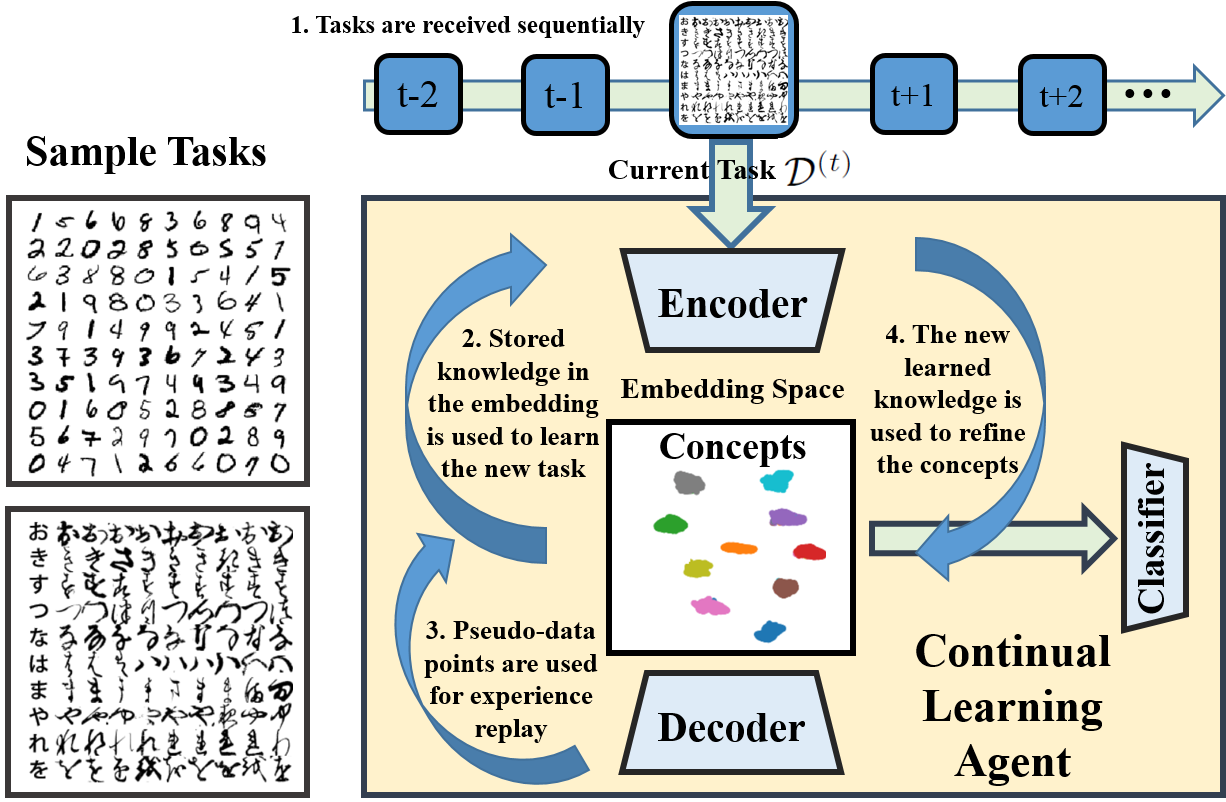}
         \caption{Architecture of the proposed framework.}
         \label{DALstructureFigCatfor}
\end{figure}

In our framework, learning concepts in each domain is considered to be classes of an ML task, e.g., different types of digit characters. We consider  a continual learning setting~\cite{ruvolo2013ella}, where an agent  receives consecutive  tasks $\{\Task{t}\}_{t=1}^{T_{\text{Max}}}$ in a sequence $t=1, \ldots, T_{\text{Max}}$ over its lifetime. The total number of tasks, distributions of the tasks,  and the order of tasks is not known a priori. 
Since the agent is a lifelong learner,  the current tasks is learned at each time step and the agent then proceeds to learn the next  task. The knowledge that is gained from   experiences is used to learn the  current task efficiently, i.e., using minimal number of labeled data. The new learned knowledge from the current task  also would be accumulated to the past experiences to potentially ease learning in future.  Additionally, this accumulation must be done consistently to generalize the learned concepts as the agent must perform well on all learned task, i.e., not to forget. This is because the learned tasks may be encountered at any time in future. Figure~\ref{DALstructureFigCatfor} presents a  high-level block-diagram
visualization of this   framework.

We model an abstract concept as a class within a domain-dependent classification task. Data points for each task $t$, are drawn i.i.d.   from the   joint probability distribution, i.e., $(\bm{x}_i^{(t)},\bm{y}_i^{(t)})\sim p^{(t)}(\bm{x},\bm{y})$ which has the  marginal distribution $q^{(t)}(\bm{x})$ over $\bm{x}$.  We consider a deep neural network $f_{\theta}:\mathbb{R}^{ d}\rightarrow \mathbb{R}^k$ as the base learning model, where $\theta$ denote the learnable weight parameters. 
A deep network is able to solve classification tasks through  extracting task-dependent high quality features in a data-driven end-to-end learning~\cite{krizhevsky2012imagenet}.
Within PDP paradigm~\cite{mcclelland1986parallel,mcclelland2003parallel,saxe2019mathematical}, this means that the data points are mapped into a discriminative embedding space, modeled by the network hidden layers, where the classes become separable, i.e., data points belonging to a class are grouped as an abstract concept. On this basis, the deep network $f_{\theta}$ is a functional composition    of  an encoder $\phi_{\bm{v}}(\cdot):\mathbb{R}^{ d}\rightarrow \mathcal{Z}\subset \mathbb{R}^f$ with learnable parameter $\bm{v}$, that encode the input data into the embedding space  $\mathcal{Z}$  and   a classifier sub-network $h_{\bm{w}}(\cdot):\mathbb{R}^{ f}\rightarrow \mathbb{R}^k$ with learnable parameters $\bm{w}$, that maps encoded information into the label space. In other words, the encoder network changes the input data  distribution as a deterministic function. Because the embedding space is discriminative, data distribution in the embedding space would be a multi-modal distribution that can be modeled as Gaussian mixture model (GMM). Figure~\ref{DALstructureFigCatfor} visualizes this intuition based on experimental data, used in the experimental validation section.

Within ML formalism, the   agent can solve the task $\Task{1}$ using standard empirical risk minimization (ERM). Given the labeled training dataset  $\mathcal{D}^{(1)} = \langle \Xt{1}, \Yt{1} \rangle$, where $\Xt{1} = [\bm{x}_1^{(1)},\ldots, \bm{x}_{n_t}^{(1)}]\in\mathbb{R}^{d\times n_1}$ and $\Yt{1}=[\bm{y}_1^{(1)},\ldots, \bm{y}_n^{(1)}]\in\mathbb{R}^{k \times n_t}$, we can solve for the network optimal weight parameters: $\hat{ \theta}^{(t)}=\arg\min_{\theta}\hat{e}_{\theta}=\arg\min_{\theta}1/n_t\sum_i \mathcal{L}_d(f_{\theta}(\bm{x}_i^{(t)}),\bm{y}_i^{(t)})$. Here, $\mathcal{L}_d(\cdot)$   is a suitable loss function, e.g., cross entropy. Conditioned on having  large enough number of labeled data points $n_1$,  the empirical risk would be a suitable   function to estimate  the real risk function, $e = \mathbb{E}_{(\bm{x},\bm{y})\sim p^{(t)}(\bm{x},\bm{y})}(\mathcal{L}_d(f_{\theta^{(t)}}(\bm{x}),\bm{y}))$~\cite{shalev2014understanding} as the Bayes optimal objective. Hence,
the trained model will generalize well on test data points for  the task $\Task{1}$. 
 Good generalization performance means that  each class would be learned as a concept which is encoded in the hidden layers. Our goal is to consolidate these learned concepts and generalize them when the next tasks with minimal 
 labeled data 
 arrive. That is, for tasks $\Task{t}, t>1$, we   have access to the dataset $\mathcal{D}^{(t)} = \langle \{\bm{X}'^{(t)}, \Yt{t}\}, \Xt{t}  \rangle$, where $\bm{X}'^{(t)} \in\mathbb{R}^{d\times n_t}$ denotes the labeled data points and $ \Xt{t}   \in\mathbb{R}^{d\times n_t}$ denotes unlabeled data points.
 This learning setting means that the learned concepts must be  generalized in the subsequent domains with minimal supervision.
 Standard ERM can not be used to learn the subsequent tasks because the number of labeled data points is not sufficient, i.e., overfitting would occur.
Additionally, even in the presence of enough labeled data, catastrophic forgetting would be consequence of using ERM. This is because the model parameters will be updated using solely the current task data which can potentially  deviate  the values of $\theta^{(T)}$ from the previous learned values  in the past time step. Hence, the agent would not retain its learned knowledge. 

Following PDP hypothesis, our goal is  to use the encoded distribution in the embedding space to expand the concepts that are captured the embedding space such that  catastrophic forgetting does not occur.   The gist of our idea is  to update the encoder sub-network such that  each subsequent task  is learned such that its distribution in the embedding space matches the distribution that is shared by  $\{\Task{t}\}_{t=1}^{T-1}$ at $t=T$. Since this distribution is initially learned via $\Task{1}$ and subsequent tasks are enforced to share this distribution in the embedding space with $\Task{1}$, we do not need to learn it from scratch as the concepts are shared across the tasks.   Hence, since the embedding 
becomes invariant with respect to any learned input task,  catastrophic forgetting would not occur.

The key challenge is to adapt the standard ERM such that the tasks share the same distribution in the embedding space becomes. To this end, we modify the base network $f_{\theta}(\cdot)$  to form a generative autoencoder by amending the model with a decoder $\psi_{\bm{u}}: \mathcal{Z}\rightarrow \mathcal{X}$.
We train the model such  the pair $(\phi_{\bm{u}},\psi_{\bm{u}})$ form an autoencoder. Doing so, we enhance the ability of the model  to encode the concepts as separable clusters in the embedding. We use the knowledge about data distribution form in the embedding to match the distributions of all tasks in the embedding. This leads to consistent generalization of the learned concepts. Additionally, since the model is generative and knowledge about past experiences is encoded in the network, we can use   CLS process~\cite{mcclelland1995there} to prevent catastrophic forgetting.
When learning a new task, pseudo-data points for the past learned tasks can be generated by sampling from the shared distribution in the embedding and feeding the samples to the decoder sub-network. These pseudo-data points are used along with new task data to learn each task.  Since   the new task is learned such that its distribution matches the past shared distribution,    pseudo-data points generated for learning future tasks would also represent the current task as well.

\section{  Proposed Algorithm} 
 Following the above framework,  learning the first task ($t=1$) reduces     to minimizing the discrimination loss   for classification and the autoencoder reconstruction loss  to solve for optimal parameters: 
\footnotesize 
 \begin{equation}
\begin{split}
\min_{\bm{v},\bm{w},\bm{u}} \mathcal{L}_{c}(\bm{X}^{(1)},\bm{Y}^{(1)}) =&\min_{\bm{v},\bm{w},\bm{u}} \frac{1}{n_1} \sum_{i=1}^{n_1}\Bigg( \mathcal{L}_d\Big(h_{\bm{w}}(\phi_{\bm{v}}\big(\bm{x}_i^{(1)})\big),\bm{y}_i^{(1)}\Big)\\&+\gamma \mathcal{L}_r\Big(\psi_{\bm{u}}\big(\phi_{\bm{v}}(\bm{x}_i^{(1)})\big),\bm{x}_i^{(1)}\Big)\Bigg),
\end{split}
\label{DALDALeq:mainSupUnsupCatfor}
\end{equation}  
\normalsize
where  $\mathcal{L}_r$ is the reconstruction loss, $\mathcal{L}_c$ is the combined loss, and $\gamma$ is a trade-off parameter.

If the base learning model is complex enough, the concepts would be formed in the embedding space as separable clusters upon learning the first task. This means that the data distribution can be modeled as a GMM distribution in the embedding. We can use standard methods such as expectation maximization to fit a GMM distribution with $k$ components   to the multimodal empirical distribution formed by the drawn samples $\{(\phi_{\bm{v}}(\bm{x}_i^{(1)}),\bm{y}_i^{(1)})_{i=1}^{n_1}\}_{i=1}^{n_1}\sim p^{(0)}$ in the embedding space.
 Let $\hat{p}_{k}^{(0)}(\bm{z})$ denote the estimated parametric GMM distribution with $k$ components. The goal is to retain this initial estimation that captures concepts when future domains are encountered. Following PDP framework, we learn the subsequent tasks such that   the current task shares the same GMM distribution with the previous learned tasks in the embedding space. We also update the estimate of the shared distribution after learning each subsequent task. Updating this distribution means generalizing the concepts to the new domains without forgetting the past domains. As a result, the distribution $\hat{p}_{J,k}^{(t-1)}(\bm{z})$ captures knowledge about past domains when  $\Task{t}$ is being learned. Moreover, we can perform experience replay by generating pseudo-data points by first drawing samples  from $\hat{p}_{J,k}^{(t-1)}(\bm{z})$  and then passing the samples through the decoder sub-network. The remaining challenge is to update the model such that each subsequent  task is learned such that its corresponding empirical distribution matches $\hat{p}_{J,k}^{(t-1)}(\bm{z})$ in the embedding space. Doing so, ensures suitability of GMM to model the empirical distribution.  
 
 To match the distributions, consider $\mathcal{D}^{(t)}_{ER} = \langle \psi(\bm{Z}_{ER}^{(t)}), \Yt{t}_{ER} \rangle$ 
 denote the pseudo-dataset for tasks $\{\Task{s}\}_{s=1}^{t-1}$, generated for experience replay when $\Task{t}$ is being learned. Following the described framework, we form the following optimization problem to learn $\Task{t}$ and generalized concepts:
\begin{equation}
\begin{split}
&\min_{\bm{v},\bm{w},\bm{u}} \mathcal{L}_{c}(\bm{X}'^{(t)},\bm{Y}^{(t)})+\mathcal{L}_{c}(\bm{X}^{T}_{(ER)},\bm{Y}^T_{(ER)})+\\&\eta D\Big(\phi_{\bm{v}}(q^{(t)}(\bm{X}^{(t)}) ),\hat{p}_{J,k}^{(t)}( \bm{Z}_{ER}^{(T)})\Big)+\\&\lambda \sum_{j=1}^k D\Big(\phi_{\bm{v}}(q^{(t)}(\bm{X}'^{(t)})|C_j),\hat{p}_{J,k}^{(t)}(\bm{Z}_{ER}^{(T)}|C_j)\Big),\hspace{3mm} \forall t\ge 2,
\end{split}
\label{DALeq:mainPrMatchCatfor}
\end{equation}    
where $D(\cdot,\cdot)$ is a suitable metric function to measure the discrepancy    between two probability distributions.  $\lambda$ and $\eta$ are a trade-off parameters. The first two terms in Eq.~\eqref{DALeq:mainPrMatchCatfor} denote the combined loss terms for each of the current task few labeled data points and the generated pseudo-dataset, defined similar to Eq.~\eqref{DALDALeq:mainSupUnsupCatfor}. The third and the fourth terms implement our idea and enforce the distribution for the current task to be close to the   distribution shared by the past learned task.  The third term is added to minimize the distance between the distribution of the current tasks and $\hat{p}_{J,k}^{(t-1)}(\bm{z})$ in the embedding space. Data labels is not needed to compute this term. The fourth term may look similar but note that we have  conditioned the distance between the two distribution on the concepts to avoid the matching challenge, i.e., when wrong concepts (or classes)  across two tasks are matched in the embedding space~\cite{globerson2006metric}. We use the few labeled data that are accessible for the current task to compute this term. Adding these terms guarantees that we can continually use  GMM   to model the shared distribution in the embedding.

 \begin{algorithm}[t]
\caption{$\mathrm{ECLA}\left (L ,\lambda, \eta \right)$\label{DALGACLalgorithm}} 
 {\small
\begin{algorithmic}[1]
\State \textbf{Input:} data $\mathcal{D}^{(1)}=(\bm{X}^{(1)},  \bm{Y}^{(t)})$.
\State\hspace{8mm} $\mathcal{D}^{(t)}=(\{\bm{X}'^{(t)},  \bm{Y}^{(t)}\},\bm{X}^{(t)})$ for $t=2,\ldots,T_\text{Max}$
\State \textbf{Concept Learning}: learning the first task ($t=1$) by solving \eqref{DALDALeq:mainSupUnsupCatfor}
\State \textbf{Fitting GMM:}
\State estimate $\hat{p}_{k}^{(0)}(\cdot)$ using $ \{\phi_{\bm{v}}(\bm{x}_i^{(1)}))\}_{i=1}^{n_t}$
\For{$t \ge 2$ }
\State \textbf{Generate the pseudo dataset:}
\State    $\mathcal{D}_{\text{ER}} = 
\{( \bm{x}_{ER,i}^{(t)}=\psi(\bm{z}_{ER,i}^{(t)}), \bm{y}_{ER,i}^{(t)})\}$
\State $(\bm{z}_{ER,i}^{(t)},\bm{y}_{ER,i}^{(t)})\sim \hat{p}_{k}^{(t-1)}(\cdot) $
\State \textbf{Update:} 
\State learnable parameters are updated  by
\State solving Eq.~\eqref{DALeq:mainPrMatchCatfor}
\State \textbf{Concept Generalization:}
\State update $\hat{p}_{k}^{(t)}(\cdot)$  using the combined samples
\State $ \{\phi_{\bm{v}}(\bm{x}_i^{(t)}),\phi_{\bm{v}}(\bm{x}_{ER,i}^{(t)})\}_{i=1}^{n_t}$
\EndFor
\end{algorithmic}}
\end{algorithm} 

The main remaining question is selection of a suitable probability distance metric $D(\cdot,\cdot)$. Common probability distance measures   such as Jensen–Shannon divergence KL divergence are not applicable for our problem as the gradient for these measures is zero when the corresponding distributions have non-overlapping supports~\cite{rabin2011wasserstein}. Since deep learning optimization problems are solved using first-order gradient-based optimization methods, we must select a distribution metric which has non-vanishing gradients.
For this reason, we select the Wasserstein Distance (WD) metric~\cite{bonnotte2013unidimensional} which satisfies this requirement and has recently been used extensively in deep learning applications to measure   the distance between two  probability distributions~\cite{courty2017optimal}. In particular,   we use Sliced Wasserstein Distance (SWD)~\cite{bonneel2015sliced} which   is a suitable approximation for WD, while it can be computed efficiently using empirical samples, drawn from two distributions. 
Our concept learning algorithm, Efficient Concept Learning Algorithm (ECLA), is summarized in Algorithm~\ref{DALGACLalgorithm}.

\section{Theoretical Analysis}
We follow a standard PAC-learning style framework to analyze our algorithm~\cite{shalev2014understanding} and using result from domain adaptation~\cite{redko2017theoretical} to demonstrate the effectiveness of our algorithm. We perform the analysis in the embedding space $\mathcal{Z}$, where the hypothesis class is the set of all the classifiers $h_{\bm{w}}(\cdot)$ parameterized by   $\bm{w}$.
  For any given model $h$ in this  class, let $e_{t}(h)$ denotes the observed risk for the domain that contains  the task $\Task{t}$, $e_{t'}(h)$ denotes the observed risk for the same model on another secondary domain, and
  $\bm{w}^*$ denotes the optimal parameter for  training the model on these two tasks jointly, i.e., $\bm{w}^*= \arg\min_{\bm{w}} e_{\mathcal{C}}(\bm{w})=\arg\min_{\bm{w}}\{ e_{t}(h)+  e_{t'}(h)\}$. 
  We  also denote the Wasserstein distance between   two given distributions as   $W(\cdot,\cdot)$. 
  We rely on the following theorem~\cite{redko2017theoretical} which relates performance of a  model trained on a particular domain to another secondary   domain.

\begin{thm}
\label{conceptlearningtheorem1}
  Consider two tasks $\Task{t}$ and $\Task{t'}$ with $n_t$ and $n_{t'}$ training data points, respectively. Let $h_{\w^{(t')}}$ be a model    trained for $\Task{t'}$, then for any $d'>d$ and $\zeta<\sqrt{2}$, there exists a constant number $N_0$ depending on $d'$ such that for any  $\xi>0$ and $\min(n_t,n_{t'})\ge N_0\max (\xi^{-(d'+2)},1)$ with probability at least $1-\xi$ for all $f_{\theta^{(t')}}$, the following holds:
\begin{equation}
\begin{split}
&e_{t}(h) - e_{t'}(h) \le\\ & W(\hat{p}^{(t)}, \hat{p}^{(t')})+ e_{\mathcal{C}}(\bm{w}^*)+ \sqrt{\big(2\log(\frac{1}{\xi})/\zeta\big)}\big(\sqrt{\frac{1}{n_t}}+\sqrt{\frac{1}{n_{t'}}}\big),
\end{split}
\label{eq:theroemfromcourtyCatForDAL}
\end{equation}  
where $\hat{p}^{(t)}$ and $\hat{p}^{(t')}$ are empirical distributions formed by the drawn samples from $p^{(t)}$ and $p^{(t')}$.
\end{thm}
 Theorem~\ref{conceptlearningtheorem1} is a broad result that provides an upper-bound on performance degradation of a trained model, when used in another domain. It suggests that if the model performs well on $\Task{t'}$ and if the upper-bound is small, then the model performs well on $\Task{t'}$. 
  The last term is a constant term which depends on the number of available samples. This term is negligible when $n_t,n_{t'}\gg 1$. The two important terms are the first and the second terms.  The first term is the Wasserstein distance between the two distributions. It may seem that according to this term, if we minimize the WD between two distributions, then the model should perform well on $\Task{t}$. But it is crucial to note that the upper-bound depends on the second term as well.
The second term suggests that the base model should be able to learn both tasks jointly. However, in the presence of  ``XOR classification problem", the tasks cannot be learned by a single model~\cite{mangal2007analysis}. This means that not only the WD between two distributions should be small, but the distributions should be aligned class-conditionally. Building upon  Theorem~\ref{conceptlearningtheorem1}, we provide the following theorem for our framework.

\begin{thm} 
\label{conceptlearningtheorem2}
Consider ECLA algorithm at learning time step $t=T$. Then all tasks $t<T$ and under the conditions of Theorem~\ref{conceptlearningtheorem1}, we can conclude:
\begin{equation}
\begin{split}
e_{t}\le & e_{T-1}^J +W(\phi(\hat{q}^{(t)}), \hat{p}_{J,k}^{(t)})+\sum_{s=t}^{T-2} W(\hat{p}_{J,k}^{(s)},\hat{p}_{J,k}^{(s+1)})\\
& +e_{\mathcal{C}}(\bm{w}^*)+\sqrt{\big(2\log(\frac{1}{\xi})/\zeta\big)}\big(\sqrt{\frac{1}{n_t}}+\sqrt{\frac{1}{n_{er,t-1}}}\big),
\end{split}
\label{eq:theroemfromcourtyCatForoursDAL}
\end{equation} 
where $e_{T-1}^J$ denotes the risk for the pseudo-task with the distribution  $\psi(\hat{p}_{J,k}^{(T-1)})$.
\end{thm}
Proof: In Theorem~\ref{conceptlearningtheorem1}, consider the task $\Task{t}$ with the  distribution $\phi(q^{(t)})$  and the pseudo-task with the distribution  $p_{k}^{(T-1)}$ in the embedding space. We can use  the triangular inequality recursively  on the term $W(\phi(\hat{q}^{(t)}), \hat{p}_{J,k}^{(T-1)})$ in Eq.~\eqref{eq:theroemfromcourtyCatForDAL}, i.e., $W(\phi(\hat{q}^{(t)}), \hat{p}_{J,k}^{(s)})\le W(\phi(\hat{q}^{(t)}), \hat{p}_{J,k}^{(s-1)})+W(\hat{p}_{J,k}^{(s)}, \hat{p}_{J,k}^{(s-1)})$ for all time steps $t \le s< T$. Adding up all the terms,  concludes Eq.~\eqref{eq:theroemfromcourtyCatForoursDAL}.

We can rely on Theorem~\ref{conceptlearningtheorem2} to demonstrate that why our algorithm can generalize concepts without forgetting the past learned knowledge. The first term in Eq.~\eqref{eq:theroemfromcourtyCatForoursDAL}  is small because, experience replay minimizes this term using the labeled pseudo-data set via ERM.  The fourth term is small since we use the few labeled data points to align the distributions class conditionally in Eq.~\eqref{DALeq:mainPrMatchCatfor}.
The last  term is a  negligible constant for $n_t, n_{er,t-1}\gg 1$. The second term denotes the distance between the task distribution and the fitted GMM. When the PDP hypothesis holds and the model learns a task well, this term is small as we can approximate  $\phi(\hat{q}^{(t)})$ with $\hat{p}_{J,k}^{(s-1)})$ (see Ashtiani et al.~\cite{ashtiani2018nearly} for a rigorous analysis of estimating a distribution with GMM). In other words, this term is small if the classes are learned as concepts.  Finally, the terms in the sum term in Eq~\ref{eq:theroemfromcourtyCatForoursDAL} are minimized because at $t=s$ we draw samples from $p_{k}^{(s-1)}$ and by learning $\psi^{-1}=\psi$ enforce that  $\hat{p}_{J,k}^{(s-1)} \approx \phi(\psi(\hat{p}_{J,k}^{(s-1)}))$.
The sum term in Eq~\ref{eq:theroemfromcourtyCatForoursDAL} models the effect of history. After learning a task and moving forward, this term potentially grows as more tasks are learned. This means that forgetting effects would increase as more subsequent tasks are learned which is intuitive. To sum up, ECLA minimizes the upper bound of $e_{t}$  in Eq~\ref{eq:theroemfromcourtyCatForoursDAL}. This means that the model can learn and remember $\Task{t}$ which in turn means that the concepts have been generalized without being forgotten on the old domains.

\section{Experimental Validation}

 \vspace{-1mm}
 We validate  our method on learning two sets of sequential learning tasks:   permuted MNIST tasks and  digit recognition tasks. These  are standard benchmark classification tasks for sequential task learning. We adjust them for our  learning setting. Each class in these tasks is considered to be a concept, and each task of the sequence is considered to be learning the concepts in a new domain.

\subsection{Learning permuted MNIST tasks}
 
Permuted MNIST tasks is standard benchmark that is designed for testing abilities of AI algorithms to overcome catastrophic forgetting~\cite{shin2017continual,kirkpatrick2017overcoming}.  The sequential tasks are generated using  the  MNIST ($\mathcal{M}$) digit recognition  dataset~\cite{lecun1990handwritten}. Each task in the sequence is generated by applying a fixed random shuffling   to the pixel values of digit images across the MNIST dataset~\cite{kirkpatrick2017overcoming}. As a result,   generated tasks are homogeneous in terms of difficulty and are suitable to perform controlled experiments. Our learning setting is different compared to prior works as we considered the case where only the data for the initial MNIST task is fully labeled. In the subsequent tasks, only few data points are labeled. 
To the best of our knowledge,  no precedent method addresses this  learning scenario for direct comparison, so we only  compared  against: a) classic back propagation (BP) single task learning, (b) full experience replay (FR) using full stored data for all the previous tasks, and (c) learning using fully labeled data (CLEER)~\cite{rostami2019Complementary}. We use the same base network structure for all the methods for fair comparison. BP is used to demonstrate that our method can address catastrophic forgetting as a lower-bound. FR is used as absolute an upper-bound to demonstrate that our method is able to learn cross-task concepts without using fully labeled data. CLEER is an instance of ECLA where fully labeled data is used to learn the subsequent tasks. We used CLEER to compare our method against an upper-bound.  

\begin{figure}[tb!]
    \centering
    \vspace{-10pt}
    \hspace{-10pt}
           \begin{subfigure}[b]{0.22\textwidth}\includegraphics[width =\textwidth,]{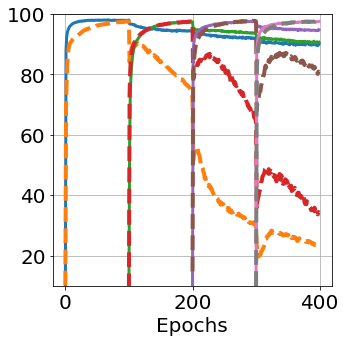}
           \centering
        \caption{BP vs. CLEER}
        \label{DALfig:BPvsCLEER}
    \end{subfigure}
    \begin{subfigure}[b]{0.22\textwidth}\includegraphics[width=\textwidth]{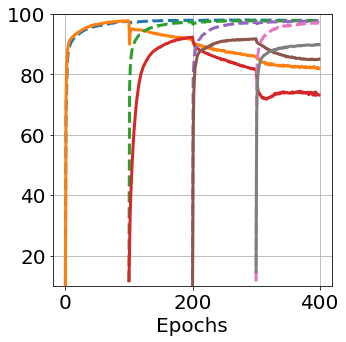}
           \centering
        \caption{ECLA vs. FR}
        \label{DALfig:ECLAvsFR}
    \end{subfigure}
      \begin{subfigure}[b]{0.22\textwidth}\includegraphics[width=\textwidth]{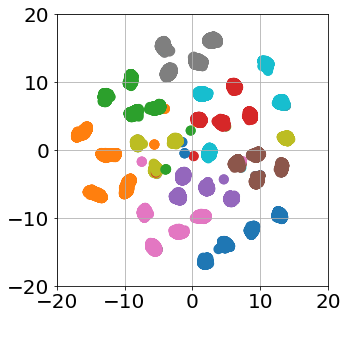}
           \centering
        \caption{ FR}
        \label{DALfig:Catfor_EWC}
    \end{subfigure}
          \begin{subfigure}[b]{0.22\textwidth}\includegraphics[width=\textwidth]{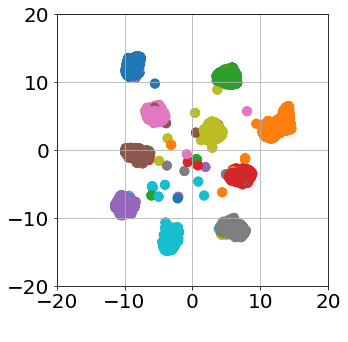}
              \centering
        \caption{ ECLA }
        \label{DALfig:Catfor_Ours}
    \end{subfigure}
     \caption{Learning curves  for four permuted MNIST   tasks((a) and (b)) and UMAP visualization of ECLA vand FR  in the embedding ((c) and (d)).  (Best viewed in color.) }\label{DALfig:resultsCatforgetPMNIST}
     \vspace{-3mm}
\end{figure}
 
  We used standard stochastic gradient descent to learn the tasks and created learning curves by computing the performance of the model on the standard testing split of the current and the past learned tasks at each learning iteration.
Figure~\ref{DALfig:resultsCatforgetPMNIST} presents  learning curves for four permuted MNIST tasks. Figure~\ref{DALfig:BPvsCLEER} presents learning curves for BP (dashed curves) and CLEER (solid curves). As can be seen, CLEER (i.e., ECLA with fully labeled data) is able to address catastrophic forgetting. This figure demonstrates that our method can be used as a new algorithm on its own to address catastrophic forgetting using experience replay~\cite{shin2017continual}.  Figure~\ref{DALfig:ECLAvsFR} presents learning curves for FR (dashed curves) and ECLA (solid  curve) when 5   labeled data points per class are used respectively. We observe that FR can tackle catastrophic forgetting perfectly but the challenge is the memory buffer requirement, which grows linearly with the number of learned tasks, making this method only suitable for comparison as an upper-bound. 
FR result also demonstrates that if we can generate high-quality pseudo-data points, catastrophic forgetting can be prevented completely. Deviation of the pseudo-data from the real data is the major reason for the initial performance degradation of ECLA on all the past learned tasks, when a new task arrives and its learning starts.
 This degradation can be ascribed to  the existing distance between $\hat{p}_{J,k}^{(T-1)}$ and $\phi(q^{(s)})$ at $t=T$ for $s < T$. Note also as our theoretical analysis predicts, the performance on a past learned task degrades more as more tasks are learned subsequently. This is compatible with the nervous system as memories fade out as time passes unless enhanced by continually experiencing a task or a concept.

\begin{figure}[tb!]
    \centering
    \vspace{-10pt}
    \hspace{-10pt}
           \begin{subfigure}[b]{0.21\textwidth}\includegraphics[width =\textwidth,]{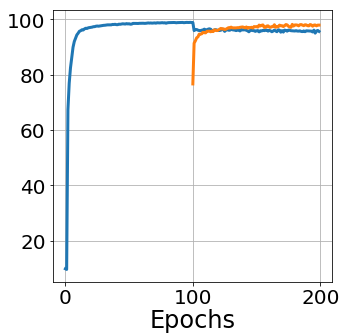}
           \centering
        \caption{$\mathcal{M}\rightarrow \mathcal{U} $}
        \label{DALfig:MNISTUSPS}
    \end{subfigure}
    \begin{subfigure}[b]{0.22\textwidth}\includegraphics[width=\textwidth]{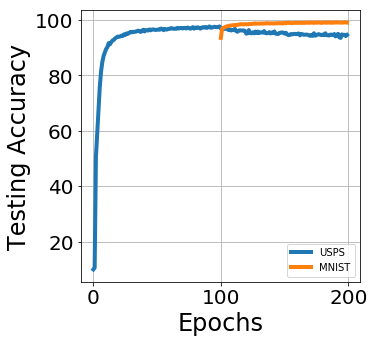}
           \centering
        \caption{$\mathcal{U}\rightarrow \mathcal{M} $}
        \label{DALfig:USPSMNIST}
    \end{subfigure}
       \begin{subfigure}[b]{0.22\textwidth}\includegraphics[width=\textwidth]{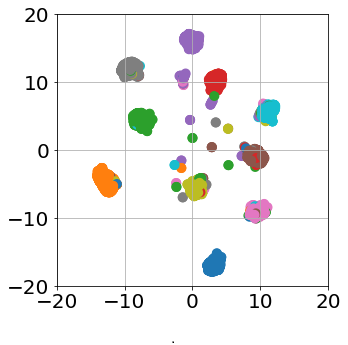}
           \centering
        \caption{$\mathcal{M}\rightarrow \mathcal{U} $}
        \label{DALfig:MNISTUSPSembed}
    \end{subfigure}
       \begin{subfigure}[b]{0.22\textwidth}\includegraphics[width=\textwidth]{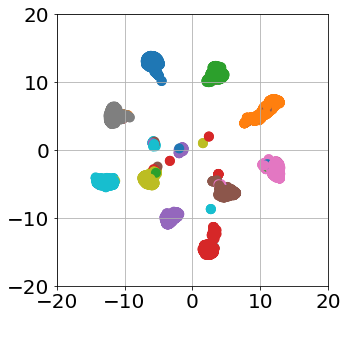}
              \centering
        \caption{$\mathcal{U}\rightarrow \mathcal{M} $}
        \label{DALfig:USPSMNISTembed}
    \end{subfigure}
     \caption{Performance results on MNIST and USPS digit recognition tasks ((a) and (b)). UMAP visualization for $\mathcal{M}\rightarrow \mathcal{U} $ and $\mathcal{U}\rightarrow \mathcal{M} $ tasks ((c) and (d)). (Best viewed in color.)}\label{DALfig:resultsCatforgetRelated}
\end{figure}
In addition to requiring  fully labeled data, we demonstrate that FR does not identify concepts across the tasks. To this end, we have visualized   the testing data  for all the tasks in the embedding space   $\mathcal{Z}$ in Figures~ \ref{DALfig:resultsCatforgetPMNIST} for FR and ECLA after learning the fourth task. For visualization purpose, we have used UMAP~\cite{mcinnes2018umap}, which  reduces the dimensionality of the embedding space to two.  In   Figure~\ref{DALfig:Catfor_EWC} and Figure~\ref{DALfig:Catfor_Ours}, each color denotes the data points of one of the digits $\{0,1,\ldots,9\}$ (each circular shape indeed is a cluster of data points). We can see that the digits form separable clusters for both methods. This result is consistent with the PDP hypothesis and is the reason behind good performance of both methods. It also demonstrates why GMM is a suitable selection to model the data distribution in the embedding space. However, we can see that when FR is used, four distinct clusters for each digit are formed (i.e., one cluster per domain for each digit class).  In other words, FR is unable to identify and generalize abstract concepts across the domains. In contrast, we have exactly ten clusters for the ten digits when ECLA is used, and hence the concepts are identified  across the domains. This is the reason that we can generalize the learned concepts to new domains, despite using few labeled data.

 \subsection{Learning  sequential digit recognition tasks}
 
  We performed a second set of experiments on a more realistic scenario. We consider two handwritten  digit recognition datasets for this purpose: MNIST ($\mathcal{M}$) and
USPS ($\mathcal{U}$) datasets.   USPS dataset is a more challenging classification task as the size of the training set is smaller (20,000 compared to 60,000 images). 
We performed experiments on    the two possible sequential learning scenarios $\mathcal{M}\rightarrow \mathcal{U} $ and $\mathcal{U}\rightarrow \mathcal{M} $.  
The experiments can be considered as    concept learning for numeral digits as both tasks are digit recognition tasks but in different domains, i.e. written by different people.

Figure~\ref{DALfig:MNISTUSPS} and Figure~\ref{DALfig:USPSMNIST} present learning curves for these two tasks when  10 labeled data points per class are used for the training of the second task. Note that the network   retains the knowledge about the first task quite well following the learning of the second task.   As expected from the theoretical justification, this empirical result suggests the performance of our algorithm depends on closeness of the distribution $\psi(\hat{p}_{J,k}^{(t)})$ to the distributions of previous tasks, and improving probability estimation will boost the performance of our approach.  Since the two tasks are much more related compared to the permuted MNIST task, forming abstract concepts is more feasible. Additionally, we see a jumststart in the performance for the second task in both Figure~\ref{DALfig:MNISTUSPS} and Figure~\ref{DALfig:USPSMNIST} which demonstrates knowledge transfer from the previous task due to the inherent similarity between the concepts.
We have also presented UMAP visualization of the data points for the tasks   in the embedding space   in Figures~\ref{DALfig:MNISTUSPSembed} and Figures~\ref{DALfig:USPSMNISTembed}.  We observe that   the distributions are matched in the embedding space and cross-domain concepts are learned by the network. These results demonstrate that our algorithm inspired by PDP and CLS theories can generalize concepts to new domains using few labeled data points. 
\begin{figure}[tb!]
    \centering
    \vspace{-10pt}
    \hspace{-10pt}
           \begin{subfigure}[b]{0.22\textwidth}\includegraphics[width =\textwidth,]{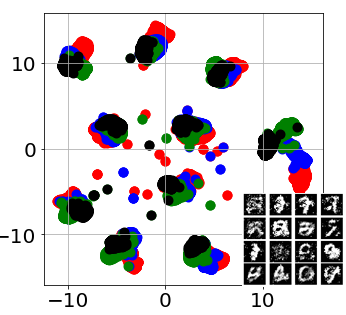}
           \centering
        \caption{Permuted MNIST tasks}
        \label{DALfig:MNISTUSPSEmbedh}
    \end{subfigure}
    \begin{subfigure}[b]{0.22\textwidth}\includegraphics[width=\textwidth]{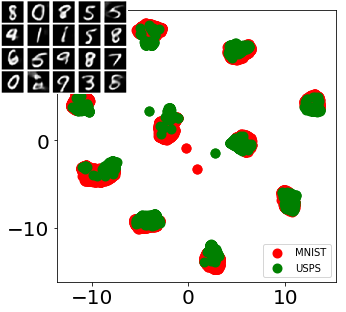}
           \centering
        \caption{$\mathcal{U}\rightarrow \mathcal{M} $}
        \label{DALfig:USPSMNISTEmbedh}
    \end{subfigure}
     \caption{UMAP visualizations of data representations for  (a) $\mathcal{U}\rightarrow \mathcal{M} $ and (b) permuted MNIST  tasks in the embedding along with a few generated pseudo-data points after learning the final task. (Best viewed in color.)}\label{DALfig:resultsCatforgetRelatedembed}
\end{figure}

 Finally, to clarify how our approach is able to learn task-agnostic concepts, Figure~\ref{DALfig:MNISTUSPSEmbedh} and Figure~\ref{DALfig:USPSMNISTEmbedh} present UMAP visualizations of data representations for  $\mathcal{U}\rightarrow \mathcal{M} $  (related and similar tasks) and permuted MNIST  tasks in the embedding, respectively. In these figures, data clusters for each task are shown with the same color. Interestingly, we observe two distinct phenomena in these two figures. When the tasks are visually similar (i.e., digit recognition tasks), data points for each class across the tasks are mixed in the embedding (in Figure~\ref{DALfig:USPSMNISTEmbedh} the green and red clusters almost completely overlap). This means that the corresponding cluster for each class in the embedding is task-agnostic. We have visualized a few random pseudo-data points that the model generated. As can be seen, with the exception of one data point,   the generated data points are similar to real digits that can represent both MNIST and USPS datasets. We conclude that when the tasks are similar, our algorithm builds clusters that represent concepts ( e.g., digit ``4'') that transcend the tasks and allow the model to generate pseudo-data points that can represent all tasks. In contrast, when the tasks are not very similar (i.e., permuted MNIST tasks), and we synthetically consider and enforce that they share the same classes, the formed clusters are structured but exhibit a different profile. In Figure~\ref{DALfig:MNISTUSPSEmbedh}, it can be seen that the data points for the four tasks in each cluster are not completely mixed; i.e., the clusters are divided among the tasks. Our algorithm works because pseudo-data points for different tasks are generated from different task-specific regions of the clusters. Of course, the model will also generate pseudo-data points that are similar to combinations of data points of two or more tasks (i.e., when we sample from a region in the cluster that is shared among several tasks), but these data points do not harm learning or cause forgetting effect as they are consistent with the clusters. This can be seen by observing the generated pseudo-data points in Figure~\ref{DALfig:MNISTUSPSEmbedh} carefully. We make an important conclusion that task-agnostic concepts can be abstracted for ``continual concept learning''  if there is an inherent similarity of the concepts across the tasks in the input space. Our algorithm also works for dissimilar tasks because we are synthetically enforcing the tasks to share the same classes, similar to the permuted MNIST tasks, but the resulting clusters would then not be completely task-agnostic. Interestingly, however, the clusters for each class across the dissimilar tasks are themselves clustered with respect to those for the other classes.
 
\section{Conclusions}
Inspired by the CLS theory and the PDP paradigm, we developed an algorithm that enables a deep network to update and generalize its learned concepts  in a continual learning setting. Our generative framework is able to encode abstract concepts in a hidden layer of the deep network in the form of a parametric GMM distribution.
This   distribution can be used to generalize concepts to new domains, where only few labeled samples are accessible. Additionally, the model is able to generate pseudo-data points for past tasks, which can be used for
  experience replay to tackle catastrophic forgetting. 
 Future work will extend our model to detect new concepts automatically and actively ask for few labeled data points as unseen concept samples are encountered.

\end{document}